\documentclass[a4paper,conference]{IEEEtran}

\ifCLASSINFOpdf
   \usepackage[pdftex]{graphicx}
   \usepackage[table,xcdraw]{xcolor}
   \graphicspath{{images/}}
   \DeclareGraphicsExtensions{.pdf,.jpeg,.png,.jpg}
\else
  
\fi
\usepackage{comment}
\usepackage{graphicx}
\usepackage{amsmath}

\ifCLASSOPTIONcompsoc
 \usepackage[caption=false,font=normalsize,labelfont=sf,textfont=sf]{subfig}
\else
 \usepackage[caption=false,font=footnotesize]{subfig}
\fi

\usepackage{url}

\usepackage{xspace}
\makeatletter
\DeclareRobustCommand\onedot{\futurelet\@let@token\@onedot}
\def\@onedot{\ifx\@let@token.\else.\null\fi\xspace}
\def\eg{\emph{e.g}\onedot}

\def\etc{\emph{etc}\onedot}

\makeatother

\usepackage{color}

\usepackage{multirow}

\usepackage{url}
\usepackage{hyperref}
\usepackage{tabularx}

\begin{document}

\title{Rethinking Domain Generalization Baselines}

\author{
\IEEEauthorblockN{Francesco Cappio Borlino$^1$,  Antonio D'Innocente$^2$$^,$$^3$, Tatiana Tommasi$^1$$^,$$^3$}
\IEEEauthorblockA{
$^1$Politecnico di Torino, Turin, Italy email: francesco.cappio@polito.it, tatiana.tommasi@polito.it \\
$^2$University of Rome Sapienza, Rome, Italy email email: dinnocente@diag.uniroma1.it \\
$^3$Italian Institute of Technology, Turin, Italy \\
}
}

\maketitle

\begin{abstract}
Despite being very powerful in standard learning settings, deep learning models can be extremely brittle when deployed in scenarios different from those on which they were trained. Domain generalization methods investigate this problem and data augmentation strategies have shown to be helpful tools to increase data variability, supporting model robustness across domains. 
In our work we focus on style transfer data augmentation and we present how it can be implemented with a simple and inexpensive strategy to improve generalization. Moreover, we analyze the behavior of current state of the art domain generalization methods when integrated with this augmentation solution: our thorough experimental evaluation shows that their original effect almost always disappears with respect to the augmented baseline. This issue open new scenarios for domain generalization research, highlighting the need of novel methods properly able to take advantage of the introduced data variability. 

\end{abstract}

\IEEEpeerreviewmaketitle

\section{Introduction}
\label{sec:intro}
The real world offers such a large diversity that the standard machine learning assumption of collecting train and test data under the same conditions, thus from the same domain/distribution, is broadly violated. Domain adaptation and domain generalization methods tackle this problem under different points of view. In the first case, unlabeled test data are considered available at training time, allowing the learning model to peek into the characteristics of the target set and adapt to it~\cite{csurka_book}. Domain generalization is a more challenging task because target data are fed to the system only during deployment~\cite{NIPS2011_blanchard,shallowDG}. In this last setting it is crucial to train robust model, possibly exploiting multiple available sources. Towards this goal, most of the existing domain generalization strategies try to incorporate the observed data invariances, capturing them at feature~\cite{Li_2018_CVPR} or model (meta-learning~\cite{episodic_hospedales} and self-supervision~\cite{lopez_rotation}) level, in the hypothesis that analogous invariances hold for future test domains. An alternative solution consists in extending the source domains by synthesizing new images. This is usually done by learning generative models with the specific constraint of preserving the object content but varying the global image appearance, with the aim of better spanning the data space and include a larger variability in the training set. Thanks to the developments in generative learning, it is becoming more and more evident that their integration into domain generalization approaches is effective~\cite{zhou2020deep}. However their performance tends to grow together with the complexity of the learning procedure which may involve one or multiple generator modules and adversarial training. We also noticed a particular trend in the most recent domain generalization research. 
Several papers discuss the merit of the proposed data augmentation solutions in comparison with feature and model-based generalization techniques~\cite{zhou2020deep,zhang2020learning}. 
Still, newly introduced feature and model-based approaches avoid benchmarks against data augmentation strategies, probably considering them unfair competitors due to the extended training set~\cite{wang2019learning,huang2020selfchallenging}. We believe that the field needs some clarification and we dedicate our work on this topic. Specifically our main contributions are:\\

\begin{figure}
\centering
\begin{tabular}{|c|cc|}
\hline
 & \multicolumn{2}{c|}{content images}\\\hline
& \includegraphics[width=0.28\linewidth]{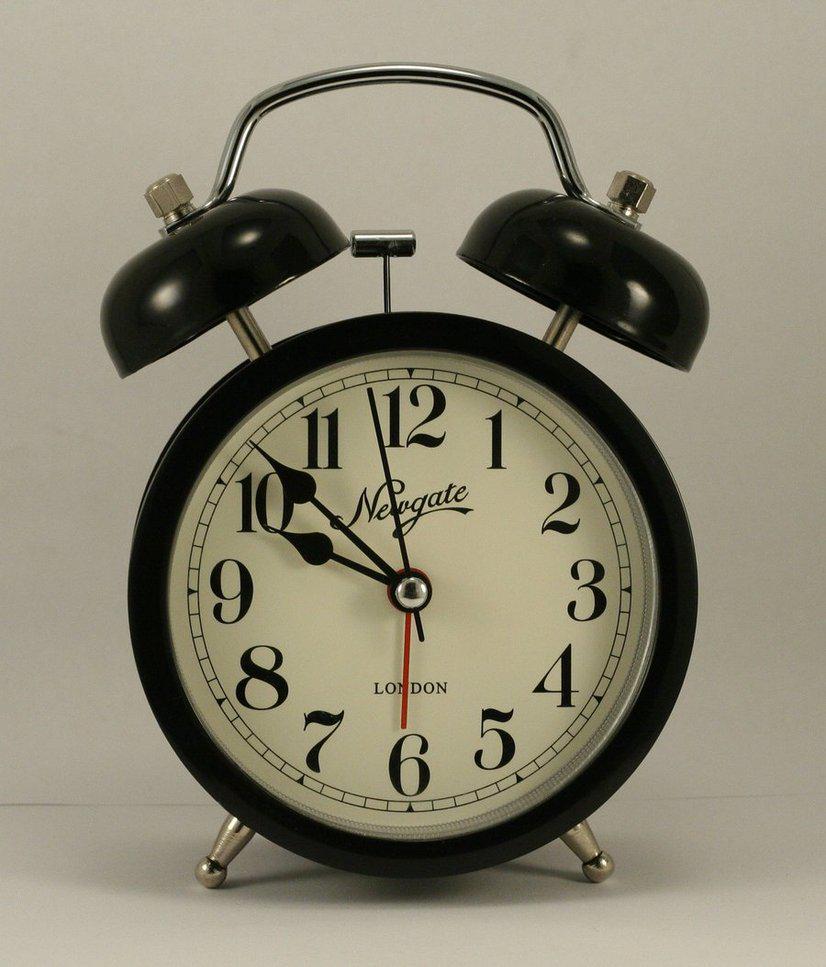} &
\includegraphics[width=0.28\linewidth]{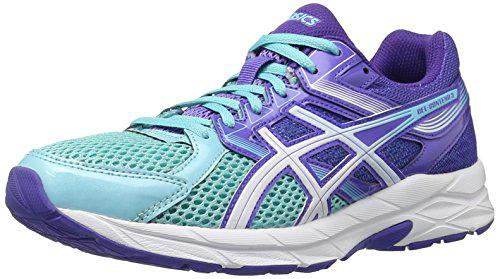}\\
\hline
style image & \multicolumn{2}{c|}{stylized images}\\ \hline
\includegraphics[width=0.24\linewidth]{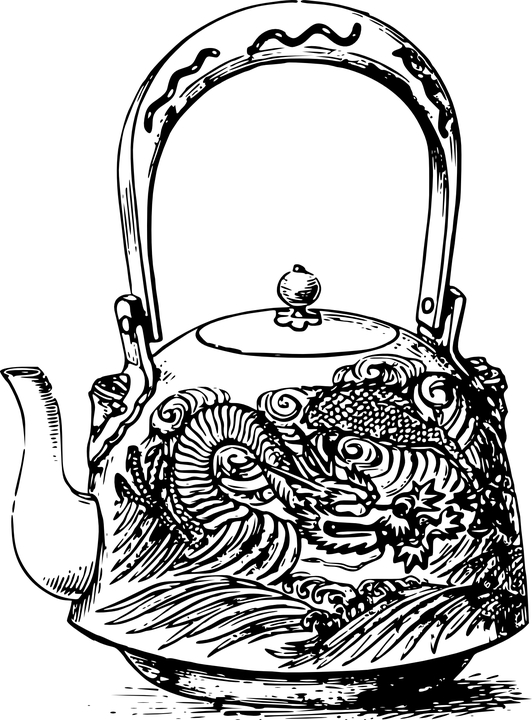} &
\includegraphics[width=0.28\linewidth]{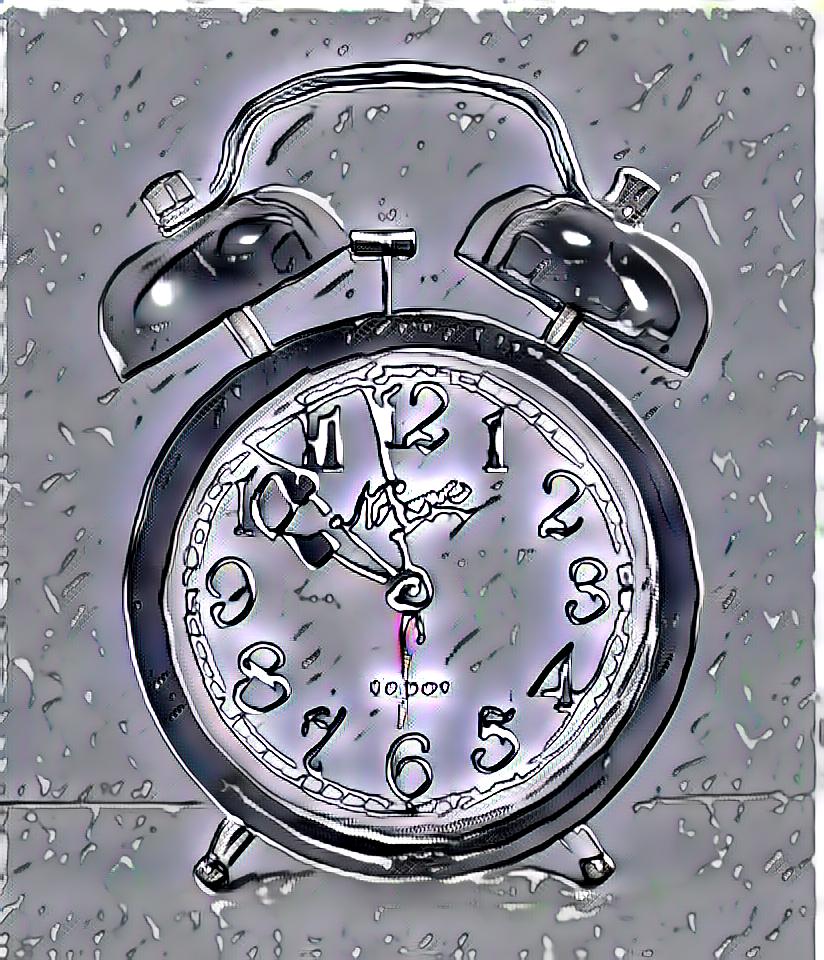} &
\includegraphics[width=0.28\linewidth]{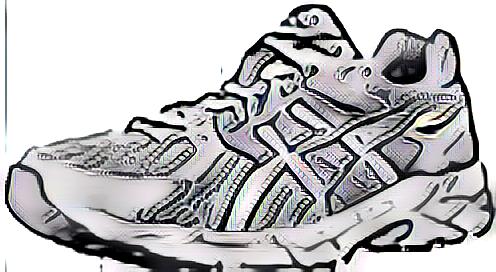} \\
\hline
\includegraphics[width=0.28\linewidth]{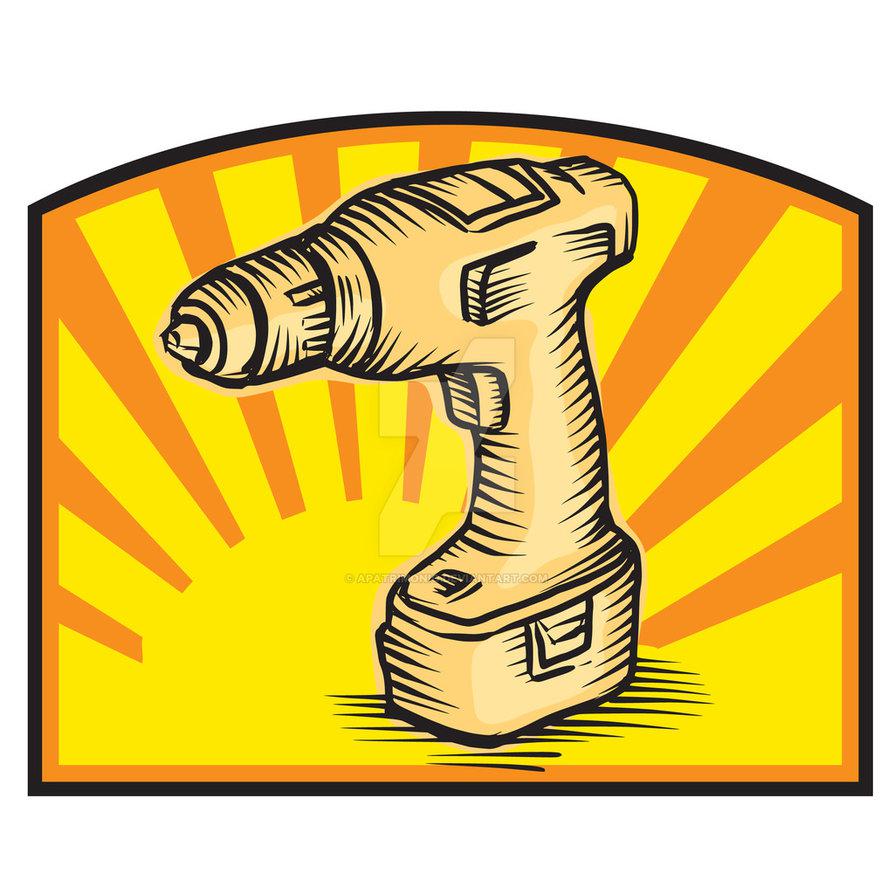} &
\includegraphics[width=0.28\linewidth]{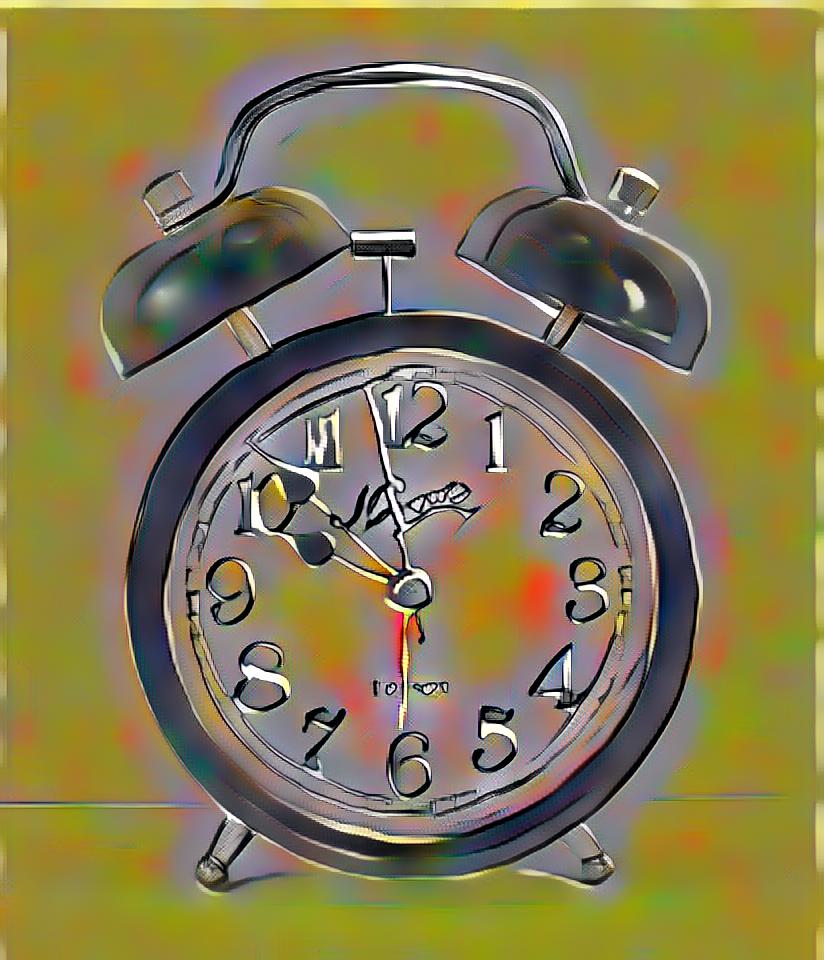} &
\includegraphics[width=0.28\linewidth]{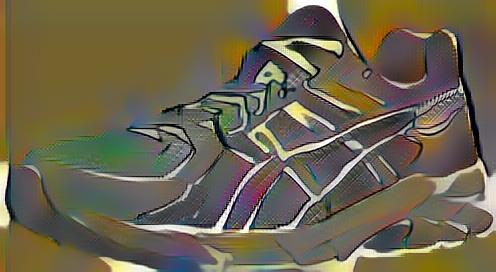} \\
\hline
\includegraphics[width=0.28\linewidth]{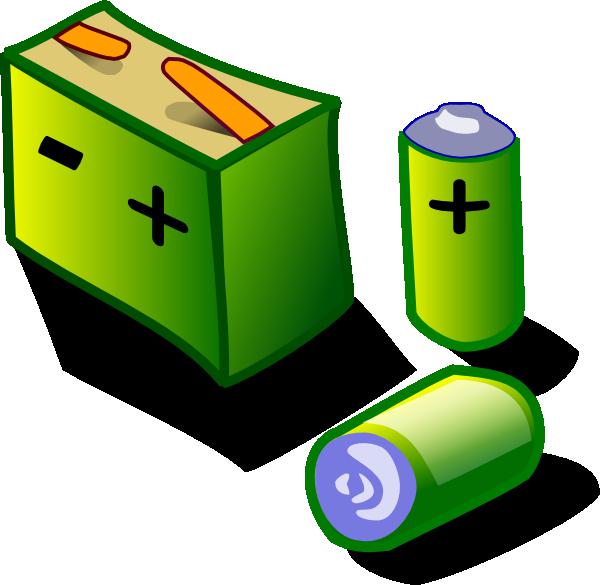} &
\includegraphics[width=0.28\linewidth]{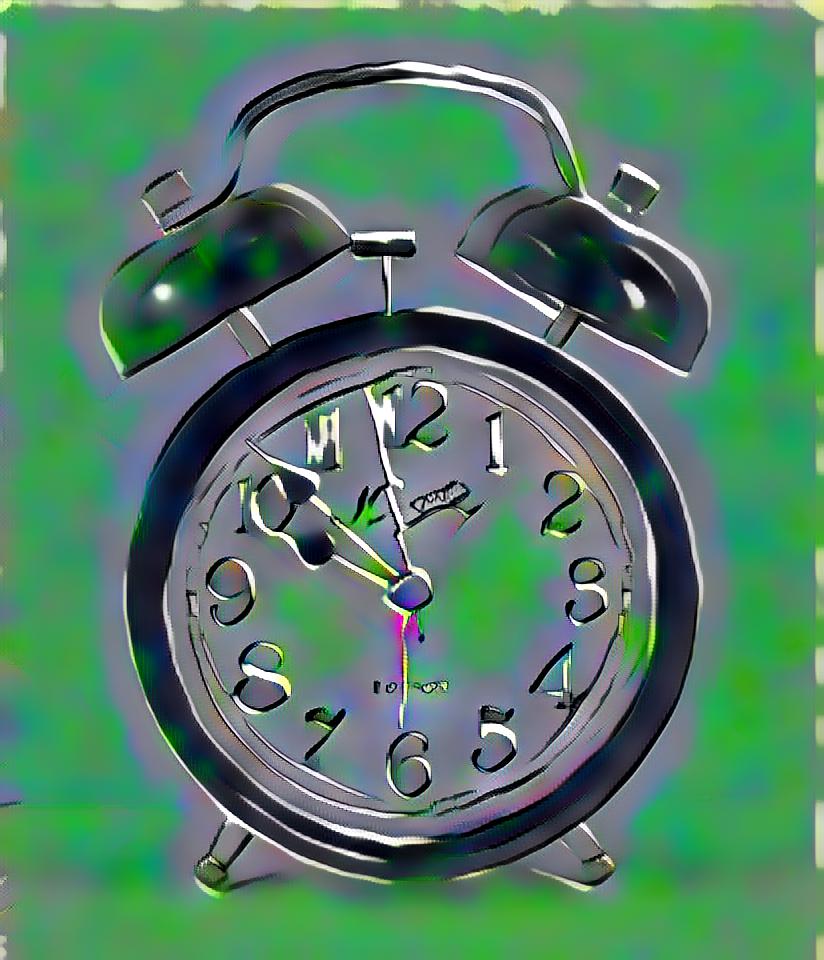} &
\includegraphics[width=0.28\linewidth]{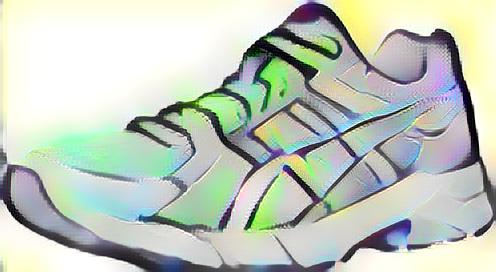} \\
\hline
\includegraphics[width=0.28\linewidth]{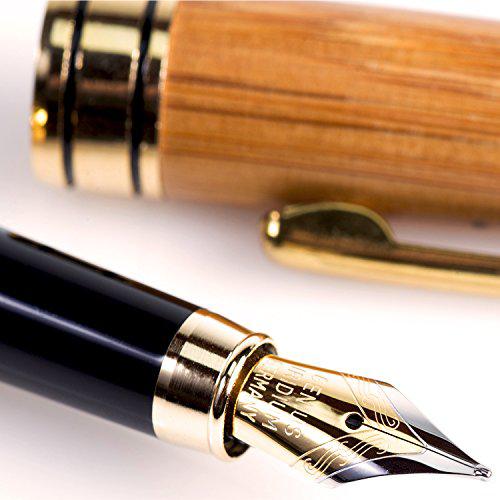} &
\includegraphics[width=0.28\linewidth]{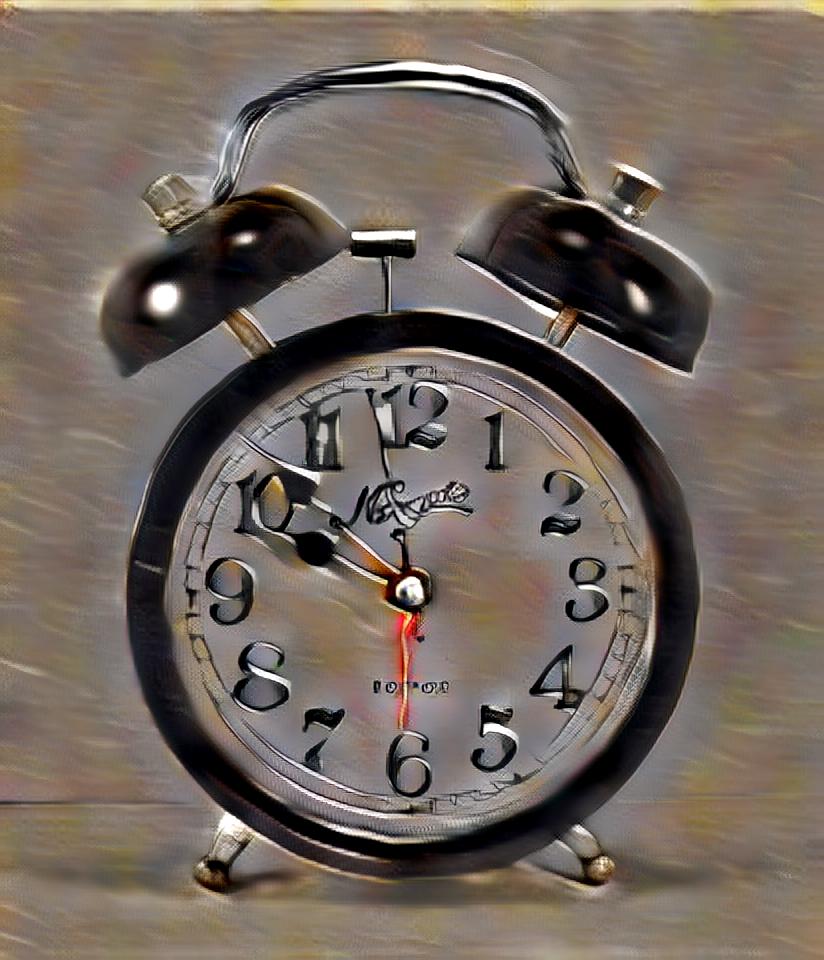} &
\includegraphics[width=0.28\linewidth]{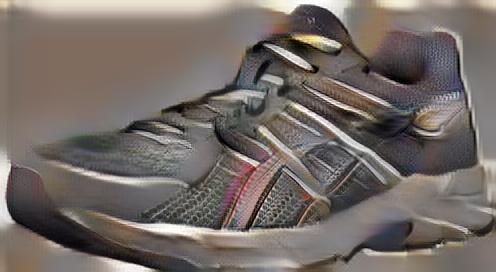} \\
\hline
\end{tabular}

\caption{Source augmentation by style transfer allows to generate different variants of each image, borrowing the style from any other image and by keeping the original semantic content. 
The images are taken from OfficeHome dataset and the style transfer is performed using AdaIN. 
}
\label{fig:examples}
\end{figure}

\vspace{-2mm}\noindent$\bullet$\hspace{2mm} \textbf{A simple and effective style transfer data augmentation approach for domain generalization}. We show how the method AdaIN~\cite{Huang_2017_ICCV_adain}, that is able to perform style transfer in real time, can be re-purposed for data augmentation, combining semantic and texture information of the available source data (see Figure \ref{fig:examples}). The extended training set allows to get top target results, outperforming existing state of the art approaches.\\

\vspace{-2mm}\noindent$\bullet$\hspace{2mm} \textbf{We designed tailored strategies to integrate for the first time style transfer data augmentation with the current state of the art approaches}. The obtained results indicate that the original advantage of those methods almost always disappears when compared with the data augmented baseline. \\

\vspace{-2mm}The scenario described by this analysis clearly suggests the need of rethinking domain generalization baselines. On one side simple data augmentation strategies should be envisaged to increase source data variability compatible with orthogonal feature and model generalization approaches. On the other, new cross-source adaptive strategies should be designed to build over images generated by style transfer approaches. 

\section{Related Work}
\label{sec:related}
The literature of domain generalization (DG) grew fast in the last years. Existing methods can be roughly divided into four main groups. 
\emph{Feature Alignment} approaches inherit the standard strategy adopted in domain adaptation which consists in measuring domain distances and learning a representation that reduce them. In the DG setting, this condition is applied among the available sources through MMD discrepancy constraints~\cite{Li_2018_CVPR} or using metric learning (contrastive loss)~\cite{doretto2017} and adversarial domain classifiers~\cite{Li_2018_ECCV}.

\emph{Meta-Learning} solutions separate the sources in meta-train and meta-test: a model is learned on the former with the real goal of reducing the error on the latter. In this way it is possible to get ready to the domain shift that will be experienced on the actual target. Two among the most well known approaches exploit episodic training with~\cite{MLDG_AAA18}, or without~\cite{episodic_hospedales} an ad hoc gradient descent update rule. Another meta-learning strategy presented in~\cite{NIPS2018_metareg} formulates a novel regularization function.

\emph{Self-supervised} learning has recently shown to support generalization. In~\cite{jigsawCVPR19} the jigsaw puzzle task was solved as auxiliary objective together with supervised object classification, helping it to focus on the object parts and their shape rather than on domain specific texture. 
A similar solution was also adopted in~\cite{dinnocente2020oneshot} using rotation recognition as side task for cross-domain detection.  
Before self-supervision, unsupervised learning already demonstrated a beneficial effect on generalization through reconstruction~\cite{Bousmalis:DSN:NIPS16} and clustering~\cite{dg_mmld} tasks. 

\emph{Data Augmentation} strategies allow to increase the source diversity: a model learned on those data gains robustness against specific features of the seen domains. 
Several approaches have been proposed to generate new samples, from the simple random changing of color or background in case of synthetic objects and robotics applications~\cite{Tobin2017DomainRF}, to the most complex use of adversarial gradients~\cite{Volpi_2018_NIPS,zhou2020deep}. Domain Mixup can also be included in the data augmentation methods~\cite{zhang2018mixup,xu2020adversarial}: pairs of examples from different domains are interpolated together with their label to learn on a more continuous domain-invariant data distribution. Finally, style transfer approaches can be used to define a specific form of data augmentation. Those methods were originally defined to match the style and content from two different images and produce a new combined visual sample. Some approaches involve complex GAN-based architectures~\cite{CycleGAN2017}, while others simply rely on data statistics and can be easily re-purposed for domain generalization~\cite{Huang_2017_ICCV_adain}.

\section{Source Augmentation by Style Transfer}
\label{sec:method}
We focus on the multi-source domain generalization setting where $S =\{S_1,\ldots,S_n\}$ denotes the $n$ available data sources with the respective $\{x^s_i,y^s_i\}_{i=1}^{N_s}\in S$ samples, where $y_i$ specifies the object classification label of its $x_i$ image. The main goal is to generalize to an unknown target database  $\{x^t_i,y^t_i\}_{i=1}^{N_t}\in T$, where $T$ shares with $S$ the same set of categories, while each source and the target are drawn from different marginal distributions. 

We indicate with $C(x^s, \theta_c)$ a basic deep learning classifier parametrized by $\theta_c$  and trained on the source data by minimizing the standard cross-entropy loss $\mathcal{L}(C(x^s, \theta_c), y^s)$. To increase data variability we study how to augment each sample $x^s$ by keeping its semantic content and changing the image style, borrowing it from the other available source data. The
stylized sample $\tilde{x}^s$ 
obtained from $x^s$ inherits its label $y^s$ and enriches the training set, possibly making the model learned by optimizing $\mathcal{L}(C(\tilde{x}^s, \theta_c), y^s)$ more robust to domain shifts.
Thus, our analysis will consider a two step process, where a deep model $A$ parametrized by $\theta_a$ is first learned on the source data to perform style transfer $x^s \rightarrow \tilde{x}^s = A(x^s, \theta_a)$, and then it is used to perform data augmentation at runtime while learning to classify the image object content. 

\subsection{Training the Style Transfer Model} 
To implement $A$ we use \emph{AdaIN} \cite{Huang_2017_ICCV_adain}, a simple and effective encoder-decoder-based approach that allows style transfer in real time. The encoder $E$ extracts representative features $f_c, f_s$ respectively from the content and the style image, the first are then re-normalized to have the same channel-wise mean and standard deviation of the second as follows:
\begin{equation}
    f_{cs}= \sigma(f_s)\left(\frac{f_c-\mu(f_c)}{\sigma(f_c)}\right) + \mu(f_s)~.
\end{equation}
Finally, the obtained feature $f_{cs}$ is mapped back to the image space through the decoder $D$ minimizing two losses: 
\begin{equation}
\mathcal{L}_{A} = \mathcal{L}_c + \lambda\mathcal{L}_s~.
\end{equation}
Both the losses measure the distance between the features re-extracted through the encoder $E(D(f_{cs}))$ from the stylized output image, and $f_{cs}$. Specifically $\mathcal{L}_c$ focuses on the content information considering the whole final feature output, while $\mathcal{L}_s$ focuses on the style information, measuring the difference of mean and standard deviation of the Relu output of several encoder layers. 

The method has two main hyperparameters $\theta_a=\{\lambda, \alpha\}$. The first controls the degree of the style transfer during training by adjusting the importance of the style loss and is generally kept fixed at $\lambda=10$. The second allows a content-style trade-off at test time by interpolating between the feature maps that are fed to the decoder with $f_{cs\alpha}=D((1-\alpha)f_c + \alpha f_{cs})$. When $\alpha=0$ the network tries to reconstruct the content image, while when $\alpha=1$ it produces the most stylized image.

\subsection{Style Transfer as Data Augmentation} 

\begin{figure*}
    \centering
    \includegraphics[width=0.9\linewidth]{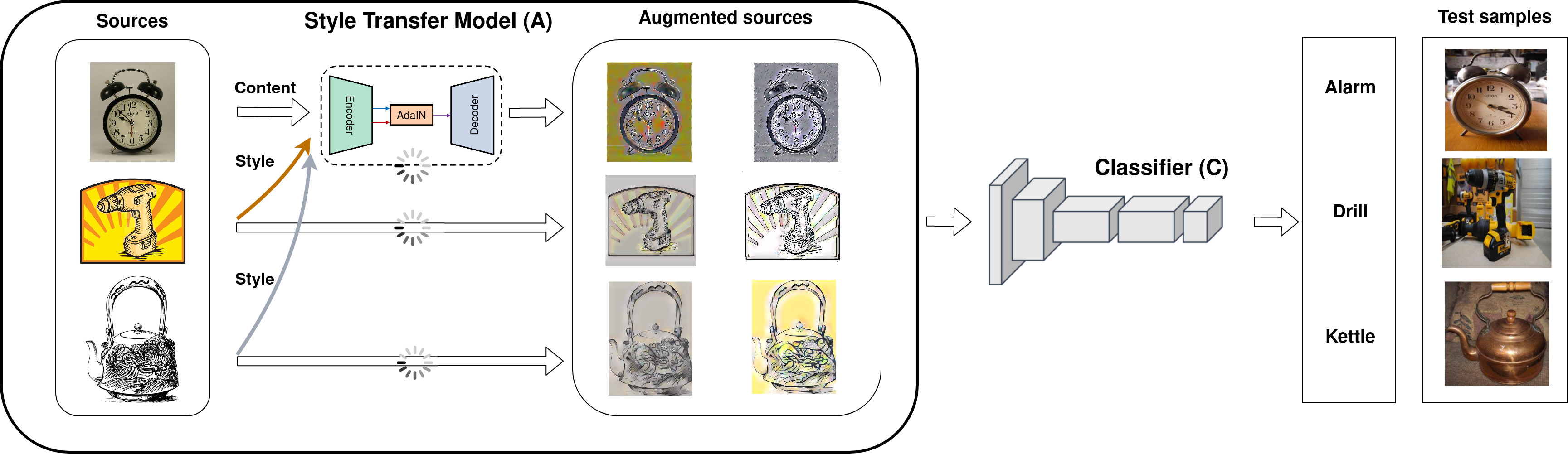}
    \caption{Classifier's training pipeline. Each training sample is augmented by borrowing the style from other images.}
    \label{fig:pipeline}
\end{figure*}

When training our object classifier $C$ the data batches contain samples extracted from all the source domains.  The samples are augmented by randomly applying the style augmentation as depicted in Figure \ref{fig:pipeline}.
Each sample in a batch has the role of content image and any of the remaining instances in the same batch can be selected randomly to work as style provider. In this scenario stylization can happen both from images of the same source domain (\eg two photos) or from images of different domains (\eg a photo and a painting). To regulate this process we use a stochastic approach with the transformed image $\tilde{x}^s$ replacing its original version $x^s$ with probability $p$.

\section{Experiments}
\label{sec:exper}
We designed our experimental analysis with the aim of running a thorough evaluation of the impact of style transfer data augmentation on domain generalization. Besides observing how this data augmentation can improve the standard learning baseline model, and how it compares with the most recent state of the art DG methods, we are also interested in the effectiveness of their combination. 
In the following we provide details on the chosen data testbeds and sota models, describing how the data augmentation strategy is integrated in each approach.

\subsection{Datasets}
We consider three standard benchmark datasets which differ in number of classes and covered domains. 

\paragraph{PACS~\cite{hospedalesPACS}} contains images of 7 object classes spanning 4 visual domains: Photo, Art Painting, Cartoon, Sketch. Given that the visual domains go from real world representations to artistic images, the style variability is quite large. We follow the original experimental protocol by training on the train splits of three source domains (using the validation splits for model selection), and then testing on the whole left out domain which acts as unknown target.  
\paragraph{OfficeHome~\cite{venkateswara2017Deep}} 
is similar to PACS, it covers 4 domains (Art, Clipart, Product and Real-World) but shows a much larger set of 65 object classes. We adopt the same experimental protocol of~\cite{Antonio_GCPR18}: a random 90-10 train-val split is used to select the training images for the 3 source domains (once again the validation images are used for model selection) and testing is performed on the whole left out target domain.
\paragraph{VLCS~\cite{TorralbaEfros_bias}} is built upon 4 different datasets: PASCAL VOC 2007, Labelme, Caltech and SUN and contains 5 object categories. Differently from the other considered testbeds, all the domains are composed of real world photos with the shift mainly due to camera type, illumination conditions, point of view,  \etc. Moreover, while Caltech is composed by object-centered images, the other three domains  contain scene images.
We apply the same experimental protocol of~\cite{jigsawCVPR19}: the predefined full training data is randomly partitioned in train and validation sets with a 90-10 ratio. The training is performed on the train splits of the 3 source domains while the validation splits are used for model selection. At the end the model is tested on the predefined test split of the left out domain. This split has been defined randomly by selecting 30\% of images of the overall dataset.

All our results are obtained by performing an average over 3 runs. In the case of both OfficeHome and VLCS the random 90-10 train-val split was repeated for each run.

\subsection{Comparison methods}
For our study we consider as main \emph{Baseline} a classification model learned on all the source data and na\"{\i}vely applied on the target. We indicate with \emph{Original} the standard data augmentation with horizontal flippling and random cropping, while we use \emph{Stylized} to specify the cases where we add style transfer data augmentation.
The behavior of four among the most recent DG methods is evaluated under both these augmentation settings. We dedicate a particular attention to the integration of the style transfer data augmentation strategy with each of the considered approaches. The goal is getting the most out of them without undermining their nature. In particular, considering that the style transfer leads to domain mixing, it is important to not integrate it in procedures that need a separation among source domains.
\paragraph{DG-MMLD~\cite{dg_mmld}} this approach exploits clustering and domain adversarial feature alignment. Since it does not need the source domain labels, the integration of the proposed style transfer data augmentation is straightforward: styles of random images are applied to each content images (inside a batch) with probability $p$, exactly as done for the Baseline.
\paragraph{Epi-FCR~\cite{episodic_hospedales}} is a meta-learning method which splits the network in two modules, each one is trained by pairing it with a partner that is badly tuned for the domain considered in the current learning episode. The modules are the feature extractor and the classifier which alternatively cover the two roles of learning part and bad reference. After this phase, a final model is learned by integrating the trained modules together with a random classifier used as regularizer. In the first stage, knowing the source domain labels is crucial to choose and set the two network modules, thus mixing the domains with style transfer augmentation could degrade its performance. In the ending stage instead, all the source data are considered together: we applied here the style data augmentation.  
\paragraph{DDAIG}~\cite{zhou2020deep} is a data augmentation strategy based on a transformation network which is trained so that every synthesized sample keeps the same label of the original image, but fools a domain classifier. In the learning procedure the transformation module, the label classifier and the domain classifier are iteratively updated. In particular the label classifier is trained on all the source data, both original and synthetic: we further extended this set with style transfer augmented data. 
\paragraph{Rotation~\cite{lopez_rotation}} it has been shown that self-supervised knowledge supports domain generalization when combined with supervised learning in a multi-task model. In particular we focused on rotation recognition, where the orientation angle of each image should be recognized among $\{0^\circ, 90^\circ, 180^\circ, 270^\circ\}$. The model minimizes a linear combination of the supervised and self-supervised loss with weight $\eta$ generally kept lower than 1 to let the supervised model guide the learning process.
In this case the domain labels are not used during training, so the application of the source augmentation by style transfer is straightforward.

An approach related to data augmentation, originally defined to improve generalization in standard in-domain learning, is \emph{Mixup}~\cite{zhang2018mixup}: it interpolates samples and their labels, regularizing a neural network to favor a simple linear behavior between training examples. Its hyper-parameter $\gamma \in \{0,\infty\}$ controls the strength of interpolation between data pairs, recovering the Baseline for $\gamma=0$. In our study we consider Mixup as further reference, and in particular we tested data mixing both at pixel and at feature level~\cite{xu2020adversarial}.

\subsection{Training setup}

Our style transfer model $A$ is trained on source data before training the classification model $C$. As already mentioned, $A$ is implemented by AdaIN~\cite{Huang_2017_ICCV_adain} and is therefore based on a VGG backbone. It is trained for 20 epochs with a learning rate equal to 5e-5. The hyperparameters $\alpha$ and $p$ used in each experiment are specified in the caption of the respective result tables and in depth analysis on the sensitivity of the method to them is presented in Section \ref{sec:sensitivity}.

For the classification model $C$ we use AlexNet and ResNet18 backbones. Specifically, \emph{Baseline}, \emph{Rotation} and \emph{Mixup} are trained using SGD with $0.9$ momentum for $30k$ iterations. We set the batch size to $32$ images per source domain: since in all the testbed there are three source domains each data batch contains $96$ images. The learning rate and the weigh decay are respectively fixed to $0.001$ and $0.0001$. Regarding the hyperparameters of the individual algorithms, we empirically set the \emph{Rotation} auxiliary weight to $\eta = 0.5$ and for \emph{Mixup} $\gamma= 0.4$.

We implement \emph{Rotation} by adding a rotation recognition branch to our Baseline.
For \emph{DG-MMLD}, \emph{Epi-FCR} and \emph{DDAIG}, we use the code provided by the authors integrating different datasets/backbones where needed. The training setup for these experiments is the one defined in their papers for both the \emph{Original} and \emph{Stylized} version. 
We report the previously published results whenever possible. In the following we will indicate with a star ($^*$) the results we obtained by running the authors' code.

\subsection{Results analysis}
\begin{table}[]
    \centering
    \caption{PACS classification accuracy (\%). We used AdaIN with $\alpha=1.0$ and $p=0.75$ for AlexNet-based experiments and AdaIN with $\alpha=1.0$ and $p=0.90$ for those based on ResNet18.} \vspace{-2mm}
    \resizebox{0.5\textwidth}{!}{
    \begin{tabular}{c@{~~}c@{~~}|c@{~~}c@{~~}c@{~~}c@{~~}|c}
    \hline
    \multicolumn{7}{c}{ AlexNet } \\
    \hline
    & & Painting & Cartoon & Sketch & Photo & Average \\
    \hline
    \multirow{5}{*}{ Original }& Baseline & 66.83 & 70.85 & 59.75 & 89.78 & 71.80 \\
    & Rotation & 65.66 & 71.89 & 62.15 & 89.88 & 72.39 \\
    & DG-MMLD & 69.27 & 72.83 & 66.44 & 88.98 & 74.38 \\
    & Epi-FCR & 64.70 & 72.30 & 65.00 & 86.10 & 72.03 \\
    & DDAIG* & 62.77 & 67.06 & 58.90 & 86.82 & 68.89 \\
    \hline
    \multirow{5}{*}{ Stylized } & Baseline & 71.96 & 72.47 & 76.47 & 88.34 & \textbf{77.31} \\
    
    & Rotation & 71.74 & 73.39 & 75.98 & 89.22 & 77.59 \\
    & DG-MMLD & 70.50 & 70.84 & 75.39 & 88.43 & 76.29 \\
    & Epi-FCR & 65.19 & 69.54 & 71.97 & 83.43 & 72.53 \\
    & DDAIG & 69.35 & 71.10 & 70.99 & 87.70 & 74.79 \\
    \hline 
    \multirow{2}{*} {Mixup}
    & pixel-level & 66.03 & 68.00 & 51.18 & 88.90 & 68.53 \\
    & feature-level & 67.04 & 69.10 & 55.40 & 88.88 & 70.11 \\
    \hline
    \multicolumn{7}{c}{ ResNet18 } \\
    \hline
    \multirow{5}{*}{ Original } & Baseline & 77.28 & 73.89 & 67.01 & 95.83 & 78.50 \\
    & Rotation & 78.16 & 76.64 & 72.20 & 95.57 & 80.64 \\
    & DG-MMLD & 81.28 & 77.16 & 72.29 & 96.06 & 81.83 \\
    & Epi-FCR & 82.10 & 77.00 & 73.00 & 93.90 & 81.50 \\
    & DDAIG* & 79.41 & 74.81 & 69.29 & 95.22 & 79.68 \\
    \hline
    \multirow{6}{*}{ Stylized } & Baseline & 82.73 & 77.97 & 81.61 & 94.95 & \textbf{84.32} \\
    & Rotation & 79.51 & 79.93 & 82.01 & 93.55 & 83.75 \\
    & DG-MMLD & 80.85 & 77.10 & 77.69 & 95.11 & 82.69 \\
    & Epi-FCR & 80.68 & 78.87 & 76.57 & 92.50 & 82.15 \\
    & DDAIG & 81.02 & 78.75 & 79.67 & 95.07 & 83.63 \\
    \hline 
    \multirow{2}{*} {Mixup}
    & pixel-level & 78.09 & 71.08 & 66.58 & 93.85 & 77.40 \\
    & feature-level  & 81.20 & 76.41 & 69.67 & 96.31 & 80.90 \\
    \hline
    \end{tabular} 
    }
    \label{tab:pacs}\vspace{-5mm}
\end{table}
\begin{table}[]
    \centering
    \caption{OfficeHome classification accuracy (\%). We used AdaIN with parameters $\alpha=1.0$ and $p=0.1$.} \vspace{-2mm}
    \resizebox{0.5\textwidth}{!}{
    \begin{tabular}{c@{~~}c@{~~}|c@{~~}c@{~~}c@{~~}c@{~~}|c}
    \hline
    \multicolumn{7}{c}{ ResNet18 } \\
    \hline
    & & Art & Clipart & Product & Real World & Average \\
    \hline
    \multirow{5}{*}{ Original } & Baseline & 57.14 & 46.96 & 73.50 & 75.72 & 63.33 \\
    & Rotation & 55.94 & 47.26 & 72.38 & 74.84 & 62.61 \\
    & DG-MMLD* & 58.08 & 49.32 & 72.91 & 74.69 & 63.75 \\
    & Epi-FCR* & 53.34 & 49.66 & 68.56 & 70.14 & 60.43 \\
    				
    & DDAIG* & 57.79 & 48.32 & 73.28 & 74.99 & 63.59 \\
    \hline
    \multirow{5}{*}{ Stylized } & Baseline & 58.71 & 52.33 & 72.95 & 75.00 & \textbf{64.75} \\
    & Rotation & 57.24 & 52.15 & 72.33 & 73.66 & 63.85 \\
    & DG-MMLD & 59.24 & 49.30 & 73.56 & 75.85 & 64.49 \\
    & Epi-FCR & 52.97 & 50.14 & 67.03 & 70.66 & 60.20 \\
    & DDAIG & 58.21 & 50.26 & 73.81 & 74.99 & 64.32 \\
    \hline
    Mixup & feature-level & 58.33 & 39.76 & 70.96 & 72.07 & 60.28 \\
    \hline
    \end{tabular} 
    }
    \label{tab:officehome} \vspace{-4mm}
\end{table}

\begin{table}[]
    \centering
    \caption{VLCS classification accuracy (\%). We used AdaIN with parameters are $\alpha=1.0$ and $p=0.75$.}\vspace{-2mm}
    \resizebox{0.5\textwidth}{!}{
    \begin{tabular}{c@{~~}c@{~~}|c@{~~}c@{~~}c@{~~}c@{~~}|c}
    \hline
    \multicolumn{7}{c}{ AlexNet } \\
    \hline
    & & CALTECH & LABELME & PASCAL & SUN & Average \\
    \hline
    \multirow{5}{*}{ Original } & Baseline & 94.89 & 59.14 & 71.31 & 64.64 & 72.49 \\
    & Rotation & 94.50 & 61.27 & 68.94 & 63.28 & 72.00 \\
    & DG-MMLD* & 96.94 & 59.10 & 68.48 & 62.06 & 71.64 \\
    & Epi-FCR* & 91.43 & 61.36 & 63.44 & 60.07 & 69.07 \\
    & DDAIG* & 95.75 & 60.18 & 65.48 & 60.78 & 70.55 \\
    \hline
    \multirow{5}{*}{ Stylized } & Baseline & 96.86 & 60.77 & 68.18 & 63.42 & 72.31 \\
    & Rotation & 96.86 & 60.77 & 68.18 & 63.42 & 72.31 \\
    & DG-MMLD & 97.49 & 61.02 & 64.23 & 62.37 & 71.28 \\
    & Epi-FCR & 92.69 & 58.18 & 62.59 & 57.87 & 67.83 \\
    & DDAIG & 97.48 & 60.48 & 65.19 & 62.57 & 71.43 \\
    \hline
    Mixup & feature-level & 94.73 & 62.15 & 69.82 & 62.98 & 72.42 \\
    \hline
    \end{tabular}
    }
    \label{tab:vlcs}\vspace{-4mm}
\end{table}
Table \ref{tab:pacs} shows results on PACS benchmark with both AlexNet and ResNet18 backbones. 
We get two main outcomes. (1) There is an evident improvement of more than 5 percentage points in the Baseline performance when using the stylized augmented source data with respect to the original case.
Looking at the results for the different domains we can see that improvement is higher for Art Painting, Cartoon and Sketch, than in Photo. 
(2) All the considered state of the art DG methods benefit from the source augmentation. Indeed in absolute terms their performance grows, but at the same time they lose in effectiveness as they cannot outperform the Baseline any more.

Table \ref{tab:officehome} shows results on OfficeHome dataset with ResNet18 backbone. Even if in this case the improvement produced by the source augmentation by style transfer is more limited, the results confirm what we have already observed for PACS. The Stylized Baseline obtains the best accuracy outperforming the competitor state of the art methods, even when those are improved using the same source augmentation.

Table \ref{tab:vlcs} reports results on VLCS benchmark with AlexNet backbone. This dataset is particularly challenging and shows a fundamental limit of tackling DG through style transfer data augmentation. Since the domain shift is not originally due to style differences in this testbed, source augmentation by style transfer does not support generalization.

As a final remark, we focus on Mixup. The results over all the considered datasets show that it is not able to generalize across domains and it might perform even worse that the Original Baseline. Between the two considered pixel and feature variants, only the second shows some advantage on PACS, so we focused on it in the other tests. Still, its results remain lower than those obtained by the DG methods both with and without style based data augmentation. 

\subsection{Analysis of AdaIN hyperparameters}
\label{sec:sensitivity}
\begin{figure}
    \centering
    \includegraphics[width=0.95\linewidth]{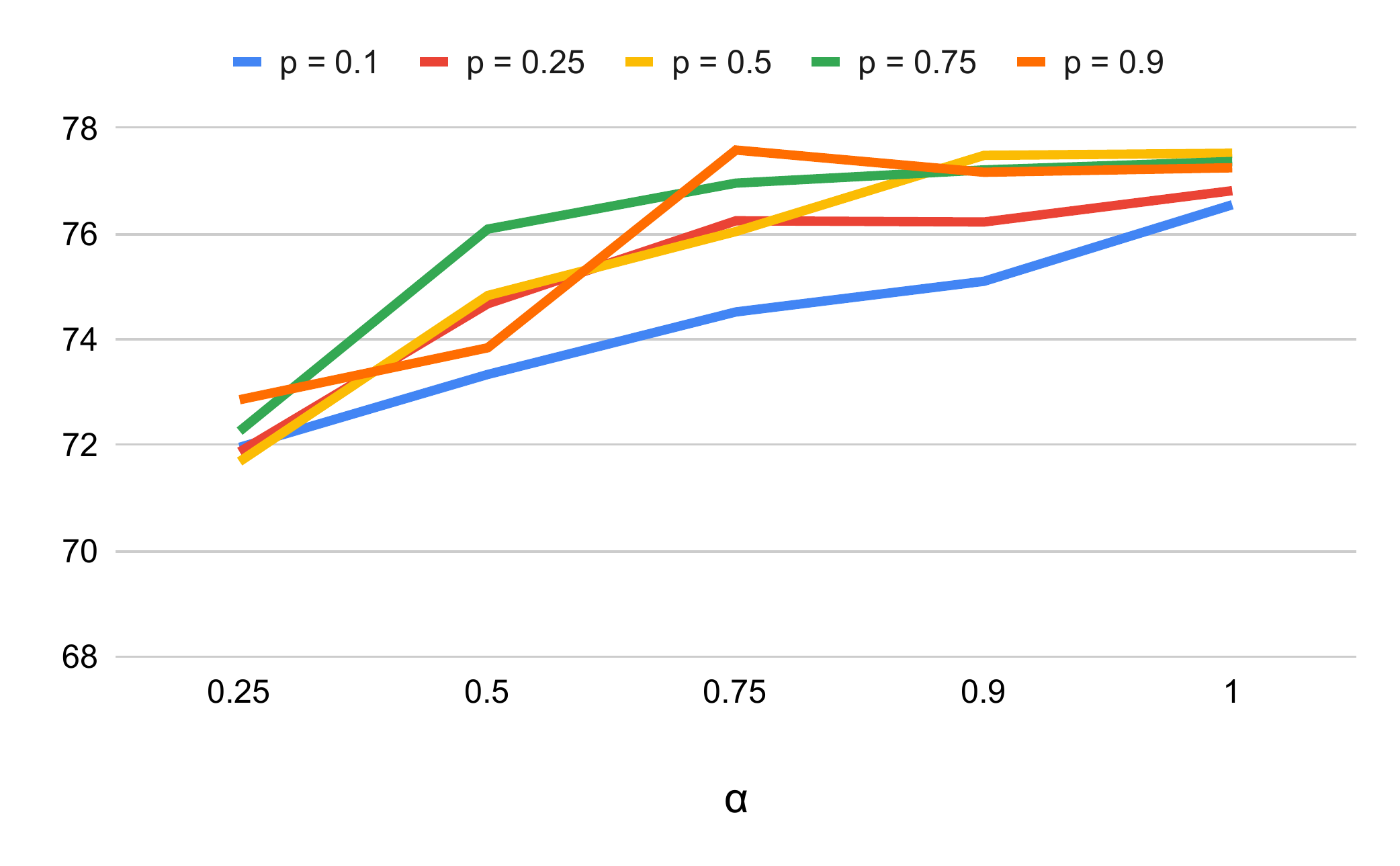}\vspace{-5mm}
    \caption{Average accuracy on PACS AlexNet with different values of $p$ when varying $\alpha$.} 
    \label{fig:varying_alpha}\vspace{-5mm}
\end{figure}

\begin{figure}
    \centering
    \includegraphics[width=0.95\linewidth]{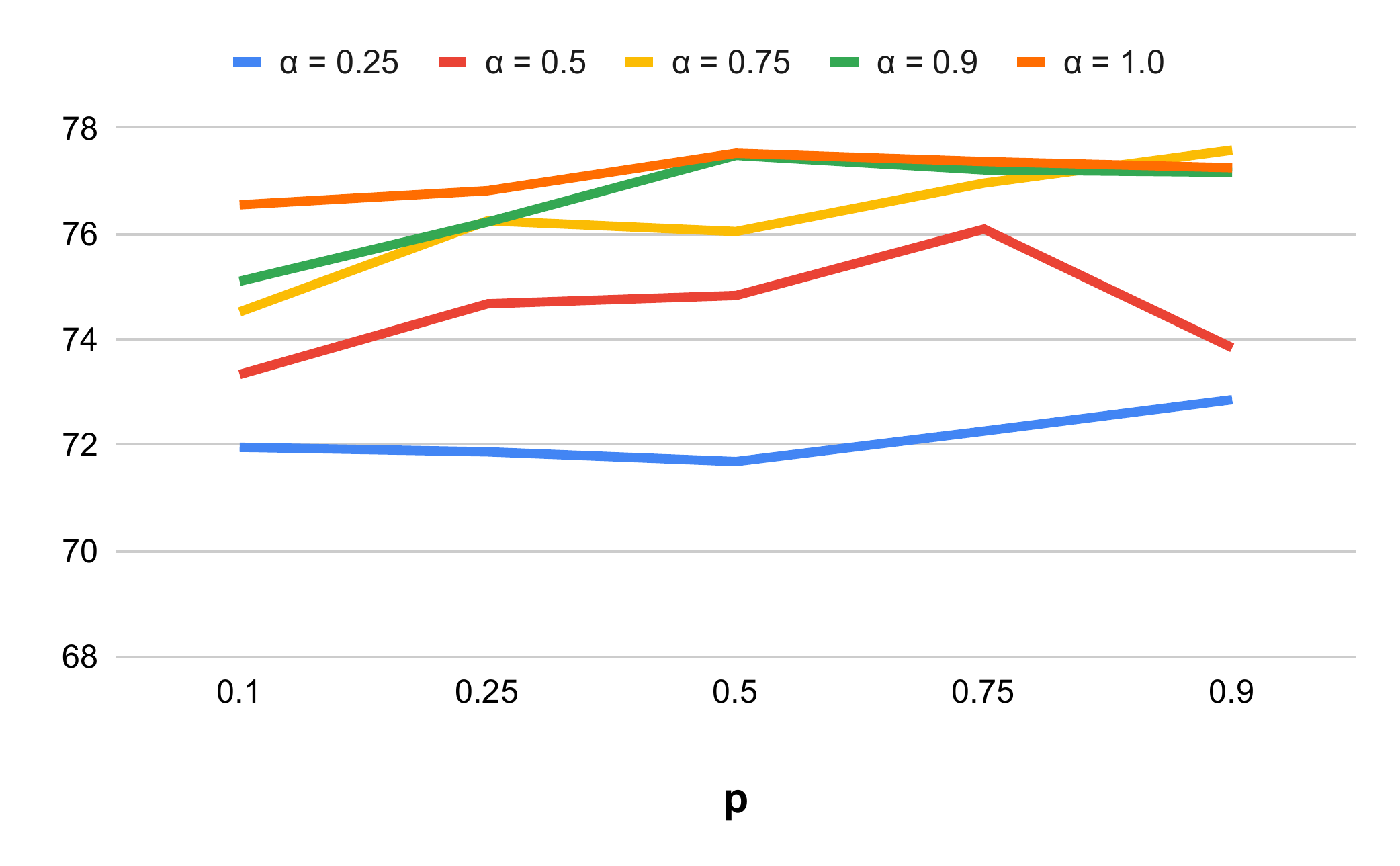}\vspace{-5mm}
    \caption{Average accuracy on PACS AlexNet with different values of $\alpha$ when varying $p$.}
    \label{fig:varying_p} \vspace{-5mm}
\end{figure}

In Figures \ref{fig:varying_alpha} and \ref{fig:varying_p} we see how the PACS AlexNet results change when varying either $\alpha$ or $p$ by keeping the other fixed. With a low value of $\alpha$ the style transfer is too weak to produce an effective appearance change of the source sample and introduce extra variability. In general the best results are obtained using $\alpha = 1$ regardless of the specific value of $p$. 

For what concerns the value of $p$ we can see that, if $\alpha$ is high enough, even a small $p$ allows to obtain good performance with the best results obtained with $p=0.5$ or $p=0.75$.


\subsection{Style transfer from external data vs source data} 
The described procedure for the application of AdaIN differs from what appeared in previous works. Indeed, both the original approach~\cite{Huang_2017_ICCV_adain} and its use for data augmentation in~\cite{zhang2020learning}, exploit the style transfer model trained on MS-COCO~\cite{mscoco} as content images, and paintings mostly collected from WikiArt~\cite{wikiart} as style images. In our study we did not allow extra datasets besides those directly involved in the domain generalization task as source domains. The reason is twofold: first, we want to keep the method as simple as possible, without the need of relying on external data; second, to perform a fair benchmark with the competitors DG methods all of them should have access to the same source information. 

\begin{figure}
\centering
\begin{tabular}{ccc}

\includegraphics[width=0.25\linewidth]{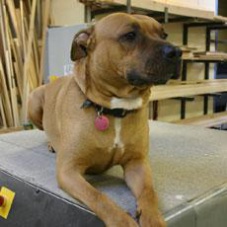} &
\includegraphics[width=0.25\linewidth]{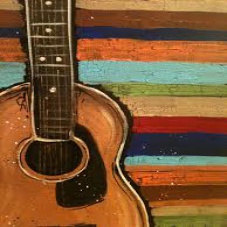} &
\includegraphics[width=0.25\linewidth]{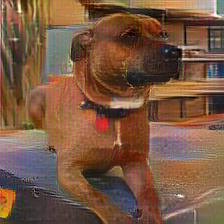}
\end{tabular}\\
\begin{tabular}{cccc}
\includegraphics[width=0.2\linewidth]{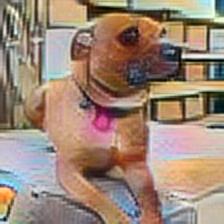} &
\includegraphics[width=0.2\linewidth]{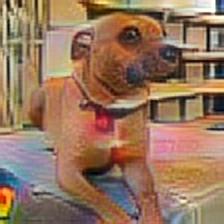} &
\includegraphics[width=0.2\linewidth]{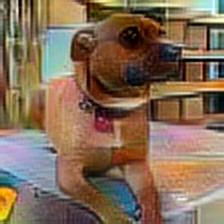} &
\includegraphics[width=0.2\linewidth]{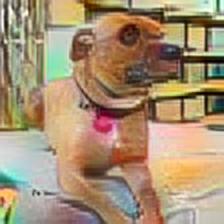}
\end{tabular}

\caption{Example of application of style transfer using AdaIN. The top left image comes from the PACS Photo domain and is used as content while the top center image comes from PACS Art Painting domain and is used as style image. On top right there is the translation performed using AdaIN trained on MS-COCO and WikiArt images. In the second row we see the translations performed using our AdaIN models trained on source data only, respectively when the Art Paintings, Cartoon, Sketch and Photo domains are used as style sources.}
\label{fig:transfer_comparison}
\end{figure}

Still, the interested reader may wonder what would be the effect of using the original AdaIN model trained on MSCOCO and WikiArt. Figure \ref{fig:transfer_comparison} shows one example obtained in this way. Specifically we consider a dog image drawn from the PACS Photo domain and we analyse the images obtained by borrowing the style form the Art Painting guitar image. 
We compare the stylized sample produced with the MSCOCO-WikiArt AdaIN model against the outcomes of the four AdaIN variants trained on the source with every one of the four domains used as target. 

As can be observed, the obtained results in terms of image quality are not so different. We also run a quantitative analysis: in Table \ref{tab:adain_training} we compare the performance of the our Stylized Baseline on PACS AlexNet with the analogous Baseline trained using the augmented data produced with the AdaIN MSCOCO-WikiArt pretrained model. The last one shows a slightly better accuracy which is though not significant if we consider the related standard deviation.

\begin{table}[tb]
    \centering
    \caption{Comparison of AdaIN training strategies}
    \resizebox{0.5\textwidth}{!}{
    \begin{tabular}{c@{~~}|c@{~~}c@{~~}c@{~~}c@{~~}|c}
    \hline
    & Art Painting & Cartoon & Sketch & Photo & Average \\
    \hline
    Stylized Baseline & $71.96$ & $72.47$ & $76.47$ & $88.34$ & $77.31 \pm 1.1$ \\
    MSCOCO-WikiArt  Baseline & $73.00$ & $73.78$ & $76.37$ & $89.04$ & $\textbf{78.05} \pm 0.9$ \\    
    \hline
    \end{tabular}
    }
    \label{tab:adain_training}
\end{table}

\section{Conclusions}
\label{sec:conclusion}
Among the current state of the art domain generalization methods some are based on data augmentation and use complex generative approaches, while other propose source feature adaptation and meta-learning strategies. Despite being orthogonal among each other, no previous work tried to integrate them. We investigated here a simple and effective style transfer data augmentation strategy for domain generalization and we showed how it overcomes its competitors. Moreover we designed proper combination of this approach with the most relevant existing DG approaches. Our experimental analysis indicates that the performance of the considered methods improves over the respective versions not including the style data augmentation, but surprisingly the methods lose their original effectiveness, not showing any improvement over the new data augmented baseline. 

As other concurrent technical reports~\cite{gulrajani2020search}, our work suggests the need of shading new light on domain generalization and calls for novel strategies able to take advantage of the data variability introduced by cross-domain style transfer.

\section*{Acknowledgment}
\noindent Computational resources provided by hpc@polito: \\ (http://hpc.polito.it).

\bibliographystyle{IEEEtran}
\bibliography{IEEEabrv,bib/root}
\end{document}